\documentclass[11pt,a4paper]{article}
\usepackage[hyperref]{emnlp-ijcnlp-2019}
\usepackage{times}
\usepackage{url}
\usepackage{latexsym}
\usepackage{amsfonts}
\usepackage{amsmath}
\usepackage{amssymb}
\usepackage{graphicx}
\usepackage{hyperref}
\newcommand{\etal}{\textit{et al. }}
\newcommand{\ie}{\textit{i}.\textit{e}.}
\newcommand{\eg}{\textit{e}.\textit{g}.}
\aclfinalcopy 

\title{Multi-Head Attention with Diversity for Learning \\Grounded Multilingual Multimodal Representations}

\author{Po-Yao Huang$^1$, Xiaojun Chang$^2$, Alexander Hauptmann$^1$  \\
  $^1$Language Technologies Institute, Carnegie Mellon University \\
  $^2$Faculty of Information Technology, Monash University \\
  \texttt{poyaoh@cs.cmu.edu, cxj273@gmail.com, alex@cs.cmu.edu}
  \texttt{}
}

\begin{document}
\maketitle
\begin{abstract}
With the aim of promoting and understanding the multilingual version of image search, we leverage visual object detection and propose a model with diverse multi-head attention to learn grounded multilingual multimodal representations.
Specifically, our model attends to different types of textual semantics in two languages and visual objects for fine-grained alignments between sentences and images.
We introduce a new objective function which explicitly encourages attention diversity to learn an improved visual-semantic embedding space.
We evaluate our model in the German-Image and English-Image matching tasks on the Multi30K dataset, and in the Semantic Textual Similarity task with the English descriptions of visual content.
Results show that our model yields a significant performance gain over other methods in all of the three tasks.
\end{abstract}

\section{Introduction}
Joint visual-semantic embeddings (VSE) are central to the success of many vision-language tasks, including cross-modal search and retrieval~\cite{KirosSZ14, karpathy2015deep, Gu_2018_CVPR}, visual question answering~\cite{vqa1,vqa2}, multimodal machine translation~\cite{mmt, elliottimagination},~etc.
Learning VSE requires extensive understanding of the content in individual modalities and an in-depth alignment strategy to associate the complementary information from multiple views.

With the availability of large-scale parallel English-Image corpora~\cite{coco,Flickr30K}, a rich line of research has advanced learning VSE under the monolingual setup. 
Most recent works~\cite{KirosSZ14, vendrov2015order, karpathy2015deep, KleinLSW15, wang2016learning, wang2018learning, what} leverage triplet ranking losses to align English sentences and images in the joint embedding space.
In VSE++~\cite{faghri2018vse++}, Faghri~\etal improve VSE by emphasizing hard negative samples.
Recent advancement in VSE models explores methods to enrich the English-Image corpora.
\newcite{shi18} propose to augment dataset with textual contrastive adversarial samples to combat adversarial attacks.
Recently, ~\newcite{ann} utilize textual semantics of regional objects and adversarial domain adaptation for learning VSE under low-resource constraints.


An emerging trend generalizes learning VSE in the multilingual scenario.
~\newcite{rajendran2016bridge} learn $M$-view representations when parallel data is available only between one pivot view and the rest of views. 
PIVOT~\cite{gella2017image} extends the work from~\newcite{calixto2017multilingual} to use images as the pivot view for learning multilingual multimodal representations.
~\newcite{lessons} further confirms the benefits of multilingual training.

\begin{figure}
    \centering
    \includegraphics[width=1.02\linewidth]{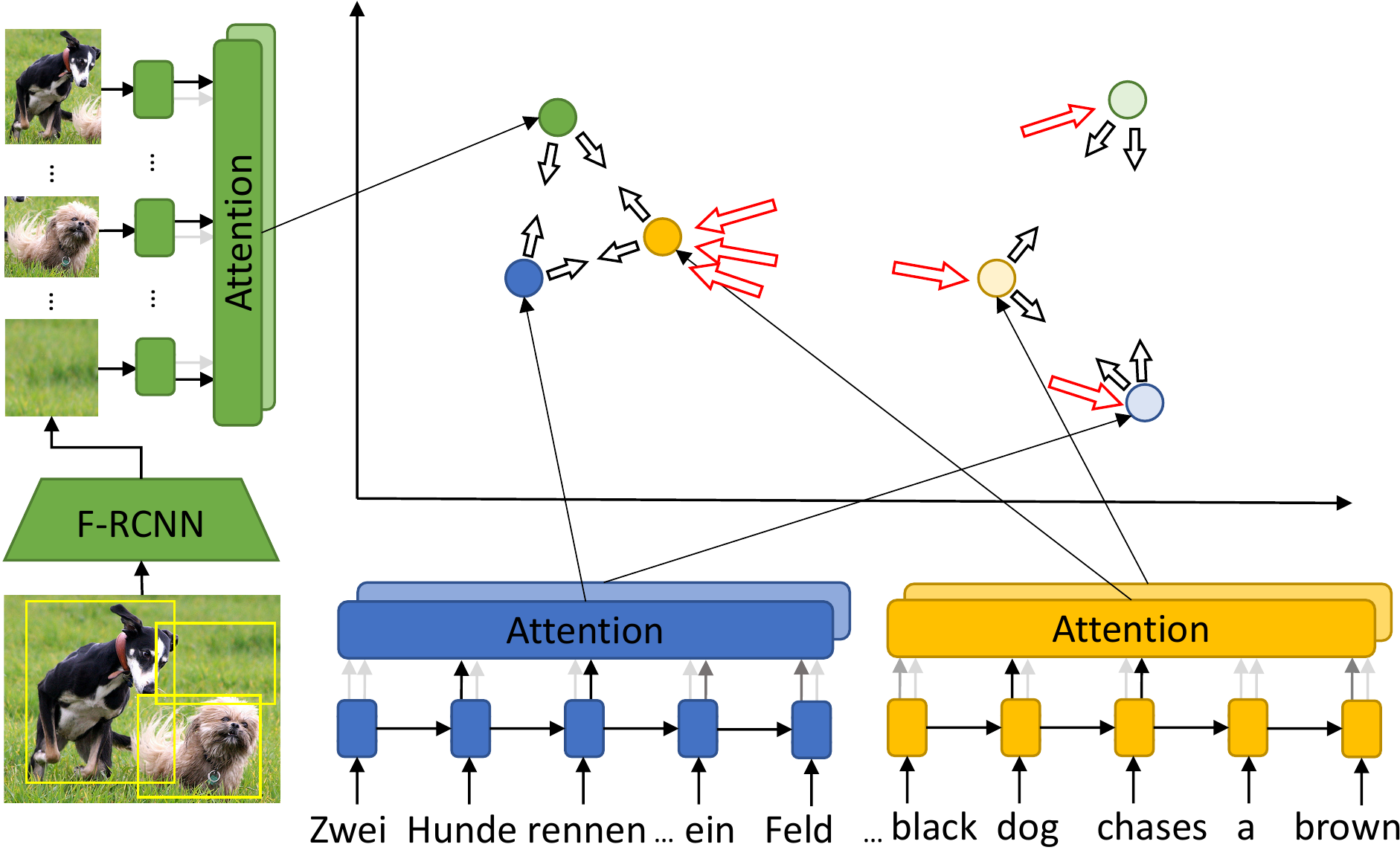}
    \caption{Multi-head attention with diversity for learning grounded multilingual multimodal representations. (A two-headed example with a part of diversity loss $l_{\theta}^D$ colored in red.) }\label{system}
\end{figure}

Our work is motivated by~\newcite{gella2017image} but has important differences. 
First, to disentangle the alignments in the joint embedding space, we employ visual object detection and multi-head attention to selectively align salient visual objects with textual phrases, resulting in visually-grounded multilingual multimodal representations.
Second, as multi-head attention~\cite{vaswani2017attention} is appealing for its efficiency and ability to jointly attend to information form different perspectives, we propose to further encourage the diversity among attention heads to learn an improved visual-semantic embedding space.
Figure~\ref{system} illustrates our gradient updates promoting diversity.
The proposed model achieves state-of-the-art performance in the multilingual sentence-image matching tasks in Multi30K~\cite{Multi30K} and the semantic textual similarity task~\cite{agirre2012semeval, agirre2014semeval,agirre2015semeval}.


\section{Related Works}

Classic attention mechanisms have been addressed for learning VSE. 
These mechanisms can be broadly categorized by the types of \{\textit{Query}, \textit{Key}, \textit{Value}\} as discussed in ~\newcite{vaswani2017attention}.
For intra-modal attention, \{\textit{Query}, \textit{Key}, \textit{Value}\} are within the same modality. 
In DAN~\cite{dan}, the content in each modalities is iteratively attended through multiple steps with intra-modal attention.
In SCAN~\cite{scan}, inter-modal attention is performed between regional visual features from~\newcite{Anderson2017up} and text semantics. The inference time complexity is $O(MN)$ (for generating $M$ query representations for a size $N$ datatset).
In contrast to the prior works, we leverage intra-modal multi-head attention, which can be easily parallelized compared to DAN and is with a preferred $O(M)$ inference time complexity compared to SCAN.

Inspired by the idea of attention regularization in~\newcite{li2018multi}, for learning VSE, we propose a new margin-based diversity loss to encourage a margin between attended outputs over multiple attention heads. 
Multi-head attention diversity within the same modality and across modalities are jointly considered in our model. 



\section{The Proposed Model}
Figure~\ref{system} illustrates the overview of the proposed model.
Given a set of images as the pivoting points with the associate English and German\footnote{For clarity in notation, we discuss only two languages. The proposed model can be intuitively generalized to more languages by summing additional terms in Eq.~\ref{loss_p} and Eq.~\ref{loss_d}-\ref{loss_all}.} descriptions or captions, the proposed VSE model aims to learn a multilingual multimodal embedding space in which the encoded representations $(v, e, g)$ of a paired instance $(i, x^e, x^g)$ are closely aligned to each other than non-paired ones.

\noindent\textbf{Encoders}: 
For a sampled $(i, x^e, x^g)$ pair, we first encode the tokens in the English sentence $x^e=\{x_1^e, \dots, x_N^e\}$ and the tokens in the German sentence $x^g=\{x_1^g, \dots, x_N^g\}$ through the word embedding matrices followed by two bi-directional LSTMs. The outputs of the textual encoders are $e=\{e_1, \dots, e_N\}, e_n \in \mathbb{R}^H$ for English and $g=\{g_1, \dots, g_N\}, g_n \in \mathbb{R}^H$ for German, 
where $N$ is the max sentence length and $H$ is the  dimension of the shared embedding space.
For the image, we leverage a Faster-RCNN~\cite{ren2015faster} network with a ResNet~\cite{he2016deep} backbone to detect and encode salient visual objects in the image. With a trainable one-layered perceptron to transform visual features into the shared embedding space, we encode the image as $v = \{v_1,\dots,v_M\}, v_m \in \mathbb{R}^H$, where $M$ is the maximum amount of visual objects in an image.

\noindent\textbf{Multi-head attention with diversity}: 
We employ $K$-head attention networks to attend to the visual objects in an image as well as the textual semantics in a sentence then generate fixed-length image/sentence representations for alignment.
Specifically, the $k$-th attended German sentence representation $g^k$ is computed by:
\begin{align}
a_{i}^k &= \text{softmax}(\text{tanh}(W^k_{c_g} c_g^k)^\top \text{tanh}(W^k_g g_i))\\  g^k &= \sum_{i=1}^{N} a_i^k g_i\label{mattn}, 
\end{align}
where $a_i^k$ is the attention weight, $W^k_g \in \mathbb{R}^{H \times H_{attn}} , W^k_{c_g} \in \mathbb{R}^{H_c \times H_{attn}}$ is the learnable transformation matrix for German. $c_g^k \in \mathbb{R}^{H_c}$ is the learnable $H_c$-dimensional contextual vector for distilling important semantics from German sentences.
The final German sentence representation is the concatenation of $K$-head attention outputs $g=[g^0\|g^1\| \dots \| g^K]$. Similar for encoding the English sentence $e=[e^0\|e^1\| \dots \| e^K]$
and the image $v=[v^0\|v^1\| \dots \| v^K]$.

With $\{V,E,G\}$ where $v \in V, e \in E, g \in G$ as the set of attended fixed-length image and sentence representations in a sampled batch, we use the widely-used hinge-based triplet ranking loss with hard negative mining ~\cite{faghri2018vse++} to align instances in the visual-semantic embedding space. Taking Image-English instances $\{V,E\}$ as an example, we leverage the triplet correlation loss defined as:
\begin{equation}
\begin{split}
    l_{\theta}(V,E) =& \sum_{p} \big[\alpha - s(v_p,e_p) + s(v_p,\hat{e}_p)\big]_+ \\
    + &\sum_{q} \big[\alpha - s(v_q,e_q) + s(\hat{v}_q,e_q)\big]_+, 
\end{split}
\end{equation}
where $\alpha$ is the correlation margin between positive and negative pairs, $[.]_+=\text{max}(0,.)$ is the hinge function, and 
$s(a,b)=\frac{a^Tb}{\|a\|\|b\|}$ is the cosine similarity. 
$p$ and $q$ are the indexes of the images and sentences in the batch. 
$\hat{e}_p=\text{argmax}_q s(v_p,e_{q \neq p})$ and $\hat{v}_q=\text{argmax}_p s(v_{p \neq q}, e_q)$ are the hard negatives.
When the triplet loss decreases, the paired images and German sentences are drawn closer down to a margin $\alpha$ than the hardest non-paired ones.
Our model aligns $\{V,E\}$, $\{V,G\}$ and $\{E,G\}$ in the joint embedding space for learning multilingual multimodal representations with the sampled $\{V,E,G\}$ batch. We formulate the overall triplet loss as:
\begin{equation}\label{loss_p}
l_{\theta}(V,E,G)= l_{\theta}(V,G) + l_{\theta}(V,E) + \gamma l_{\theta}(G,E).
\end{equation}
Note that the hyper-parameter $\gamma$ controls the contribution of $l_\theta(G,E)$ since ($e$, $g$) may not be a translation pair even though  ($e$, $v$) and  ($g$, $v$) are image-caption pairs.

\begin{table*}[t]
\centering
\scriptsize
\small
\setlength\tabcolsep{4.5pt}
\begin{tabular}{lcccccccccccc}
\hline
Method & \multicolumn{3}{c}{German to Image} & \multicolumn{3}{c}{Image to German} &\multicolumn{3}{c}{English to Image} & \multicolumn{3}{c}{Image to English}\\
& R@1 & R@5 & R10 & R@1 & R@5 & \multicolumn{1}{c}{R@10}& R@1 & R@5 & R@10 & R@1 & R@5 & R10\\ \hline\hline
VSE\textsuperscript{$\dagger$*}~\cite{KirosSZ14} & 20.3 & 47.2 & 60.1 & 29.3 & 58.1 & \multicolumn{1}{c|}{71.8} & 23.3 & 53.6 & 65.8 & 31.6 & 60.4 & 72.7 \\
OE\textsuperscript{$\dagger$*}~\cite{vendrov2015order} & 21.0 & 48.5 & 60.4 & 26.8 & 57.5 & \multicolumn{1}{c|}{70.9} & 25.8 & 56.5 & 67.8 & 34.8 & 63.7 & 74.8 \\ 

DAN\textsuperscript{*}~\cite{dan} & 31.0 & 60.9 & 71.0 & 46.5 & 77.5 & \multicolumn{1}{c|}{83.0} & 39.4 & 69.2 & 69.1 & 55.0 & 81.8 & 89.0 \\ 

VSE++\textsuperscript{*}~\cite{faghri2018vse++} & 31.3 & 62.2 & 70.9 & 47.5 & 78.5 & \multicolumn{1}{c|}{84.5} & 39.6 & 69.1 & 79.8 & 53.1 & 82.1 & 87.5 \\ 

SCAN\textsuperscript{*}~\cite{scan} & 35.7 & 64.9 & 74.6 & 52.3 & 81.8 & \multicolumn{1}{c|}{88.5} & 45.8 & 74.4 & 83.0 & 61.8 & 87.5 & 93.7 \\

\hline
Pivot\textsuperscript{$\dagger$}~\cite{gella2017image} & 22.5 & 49.3 & 61.7 & 28.2 & 61.9 & 73.4 & 26.2 & 56.4 & 68.4 & 33.8 & 62.8 & 75.2 \\ \hline
Ours\textsuperscript{$\dagger$} (Random, VGG19) & 25.8 & 54.9 & 65.1 & 34.1 & 65.5 & 76.5 & 30.1 & 62.5 & 71.6 & 36.4 & 68.0 & 80.9 \\  
Ours (Random, No diversity) & 36.3 & 65.3 & 74.7 & 53.1 & 82.3 & 88.8 & 46.2 & 74.7 & 82.9 & 63.3 & 87.0 & 93.3 \\ 
Ours (Random) & 39.2 & 67.5 & 76.7 & 55.0 & 84.7 & 91.2 & 48.7 & 77.2 & 85.0 & 66.4 & 88.3 & 93.4 \\ 
Ours (w/ FastText) & 40.3 & 70.1 & \textbf{79.0} & \textbf{60.4} & \textbf{85.4} & \textbf{92.0} & \textbf{50.1} & 78.1 & 85.7 & \textbf{68.0} & 88.8 & 94.0 \\
Ours (w/ BERT) & \textbf{40.7} & \textbf{70.5} & 78.8 & 56.5 & 84.6 & 91.3 & 48.9 & \textbf{78.3} & \textbf{85.8} & 66.5 & \textbf{89.1} & \textbf{94.1} \\ \hline
\end{tabular}
\caption{Comparison of multilingual sentence-image retrieval/matching (German-Image)  and (English-Image) results on Multi30K. (Visual encoders:VGG\textsuperscript{$\dagger$} otherwise ResNet or Faster-RCNN(ResNet).) (Monolingual models\textsuperscript{*}.) } \label{mmt_compare_Multi30}
\vspace{-1.0em}
\end{table*}

\begin{table}[]
\small
\setlength\tabcolsep{4.5pt}
\begin{tabular}{lccc}\hline
Method & MS-Vid & Pascal  & Pascal \\ 
& (2012) & (2014) & (2015) \\\hline\hline
STS Baseline & 29.9  & 51.3 & 60.4 \\
STS Best System & 86.3  & 83.4 & 86.4 \\
GRAN~\cite{wieting2017learning} & 83.7 & 84.5 & 85.0 \\ \hline
VSE~\cite{KirosSZ14} & 80.6 & 82.7 & 89.6 \\
OE~\cite{vendrov2015order} & 82.2 & 84.1 & 90.8 \\
DAN~\cite{dan} & 84.1 & 84.3 & 90.8 \\
VSE++~\cite{faghri2018vse++} & 84.5 & 84.8 & 91.2 \\
SCAN~\cite{scan} & 84.0 & 83.9 & 90.7 \\ 
PIVOT~\cite{gella2017image} & 84.6 & 84.5 & 91.5 \\ \hline
Ours (w/ Random) & 85.8 & 87.8 & 91.5 \\
Ours (w/ FastText) & 86.2 & \textbf{88.3} & \textbf{91.8} \\
Ours (w/ BERT) & \textbf{86.4} & 88.0 & 91.7 \\\hline
\end{tabular}
\caption{Results on the image and video datasets of Semantic Textual Similarity task.
 (Pearson'’s $r \times 100$ ).}\label{result_sts}
\vspace{-1.0em}
\end{table}

One of the desired properties of multi-head attention is its
ability to jointly attend to and encode different information in the embedding space. 
However, there is no mechanism to support that these attention heads indeed capture diverse information.
To encourage the diversity among $K$ attention heads for instances within and across modalities, we propose a new simple yet effective margin-based diversity loss. 
As an example, the multi-head attention diversity loss between the sampled images and the English sentences (\ie~  diversity across-modalities) is defined as:
\begin{equation}
l^{D}_{\theta}(V,E)=\sum_{p} \sum_k \sum_{r} \big[\alpha_{D} - s(v_p^k,e_p^{k \neq r})]_+ \\
\end{equation}

As illustrated with the red arrows for update in Figure~\ref{system}, the merit behind this diversity objective is to increase the distance (up to a diversity margin $\alpha_D$) between attended embeddings from different attention heads for an instance itself or its cross-modal parallel instances. 
As a result, the diversity objective explicitly encourage multi-head attention to concentrate on different aspects of information sparsely located in the joint embedding space to promote fine-grained alignments between multilingual textual semantics and visual objects.
With the fact that the shared embedding space is multilingual and multimodal,
for improving both intra-modal/lingual and inter-modal/lingual diversity, we model the overall diversity loss as:
\begin{equation}
\begin{split}\label{loss_d}
l^{D}_{\theta}(V,E,G)&= l^{D}_{\theta}(V,V) + l^{D}_{\theta}(G,G) + l^{D}_{\theta}(E,E) \\
&+ l^{D}_{\theta}(V,E) + l^{D}_{\theta}(V,G) + l^{D}_{\theta}(G,E),
\end{split}
\end{equation}
where the first three terms are intra-modal/lingual and the rest are cross-modal/lingual. With Eq.~\ref{loss_p} and Eq.~\ref{loss_d},
we formalize the final model loss as:
\begin{equation}
l_{\theta}^{All}(V,E,G)=l_{\theta}(V,E,G)+\beta l^{D}_{\theta}(V,E,G),~\label{loss_all}
\end{equation}
where $\beta$ is the weighting parameter which balances the diversity loss and the triple ranking loss. We train the model by minimizing $l_{\theta}^{All}(V,E,G)$.

\section{Experiments}
Following~\newcite{gella2017image}, we evaluate on the multilingual sentence-image matching tasks in Multi30K~\cite{Multi30K} and the semantic textual similarity task~\cite{agirre2012semeval,agirre2015semeval}.

\subsection{Experiment Setup}
We use the model in~\newcite{Anderson2017up} which is a Faster-RCNN~\cite{ren2015faster} network pre-trained on the MS-COCO~\cite{coco} dataset and fine-tuned on the Visual Genome~\cite{genome} dataset to detect salient visual objects and extract their corresponding features. 1,600 types of objects are detectable.
We then pack and represent each image as a $36\times2048$ feature matrix where 36 is the maximum amount of salient visual objects in an image and 2048 is the dimension of the flattened last pooling layer in the ResNet~\cite{he2016deep} backbone of Faster-RCNN.

For the text processing, we lower-case, tokenize, and then truncate the maximum sentence length to 100. 
We use 300-dim word embedding matrices initialized either randomly or with pre-trained multilingual embeddings. (We use the multilingual version of FastText~\cite{fast}). 
We also experiment incorporating the last layer of contextualized multilingual BERT embeddings~\cite{bert} to replace the word embedding matrices as the textual input features for the bi-directional LSTMs.

For training, we sample batches of size 128 and train 20 epochs on the training set of Multi30K. We use the Adam~\cite{kingma2014adam} optimizer with $2\times10^{-4}$ learning rate then $2\times10^{-5}$ after 15-\textit{th} epoch. Models with the best summation of validation R@1,5,10 are selected to generate image and sentence embeddings for testing.
Weight decay is set to $10^{-6}$ and gradients larger than $2.0$ are clipped. 
We use 3-head attention ($K=3$) and the embedding dimension $H=512$.
The same dimension is shared by all the context vectors in the attention modules. 
Other hyper-parameters are set as follows: $\alpha=0.2, \alpha_{D}=0.1, \beta=1.0$ and $\gamma=0.6$. 


\subsection{Multilingual Sentence-Image Matching}
We evaluate the proposed model in the multilingual sentence-image matching (retrieval) tasks on Multi30K: (i) Searching images with text queries (Sentence-to-Image). (ii) Ranking descriptions with image queries (Image-to-Sentence). English and German are considered.

Multi30K~\cite{Multi30K} is the multilingual extension of Flickr30K~\cite{Flickr30K}. 
The training, validation, and testing split contain 29K, 1K, and 1K images respectively.
Two types of annotations are available in Multi30K: (i) One parallel English-German translation for each image and (ii) five independently collected English and five German descriptions/captions for each image. We use the later. Note that the German and English descriptions are \textit{not} translations of each other and may describe an image differently.

As other prior works, we use recall at $k$ (R@$k$) to measure the standard ranking-based retrieval performance. Given a query, R@$k$ calculates the percentage of test instances for which the correct one can be found in the top-$k$ retrieved instances. Higher R@$k$ is preferred.

Table~\ref{mmt_compare_Multi30} presents the results on the Multi30K testing set. The VSE baselines in the first five rows are trained with English and German descriptions independently. In contrast, PIVOT~\cite{gella2017image} and the proposed model are capable of handling multilingual input queries with single model. For a fair comparison with PIVOT, we also report the result of swapping Faster-RCNN with VGG as the visual feature encoder in our model.

As can be seen, the proposed models successfully obtain state-of-the-art results, outperforming other baselines by a significant margin. 
German-Image matching benefit more from joint training with English-Image pairs. 
The models with pre-trained multilingual embeddings and contextualized embeddings achieve better performance in comparison to randomly initialized word embeddings, especially for German. 
One explanation is that the degradation from German singletons is alleviated by the multi-task training with English and the pre-trained embeddings. 
While the model with BERT performs better in English, FastText is preferred for German-Image matching.

\vspace{-0.2em}
\subsection{Semantic Textual Similarity Results}
\vspace{-0.2em}
For semantic textual similarity (STS) tasks, 
we evaluate on the video task from STS-2012 ~\cite{agirre2012semeval} and the image
tasks from STS-2014-15~\cite{agirre2014semeval, agirre2015semeval}.
The video descriptions are from the MSR video description corpus~\cite{chen2011collecting} and the image descriptions are from the PASCAL dataset~\cite{rashtchian2010collecting}. 
In STS, a system takes two input sentences and output a semantic similarity ranging from [0,5].
Following~\newcite{gella2017image}, we directly use the model trained on Multi30K to generate sentence embeddings then scaled the cosine similarity between the two embeddings as the prediction.

Table~\ref{result_sts} lists the standard Pearson correlation coefficients $r$ between the system predictions and the STS gold-standard scores. We report the best scores achieved by paraphrastic embeddings~\cite{wieting2017learning} (text only) and the VSE models in the previous section. Note that the compared VSE models are all with RNN as the text encoder and \textit{no} STS data is used for training. Our models achieve the best performance and the pre-trained word embeddings are preferred.

\begin{figure}
    \centering
    \includegraphics[width=0.95\linewidth]{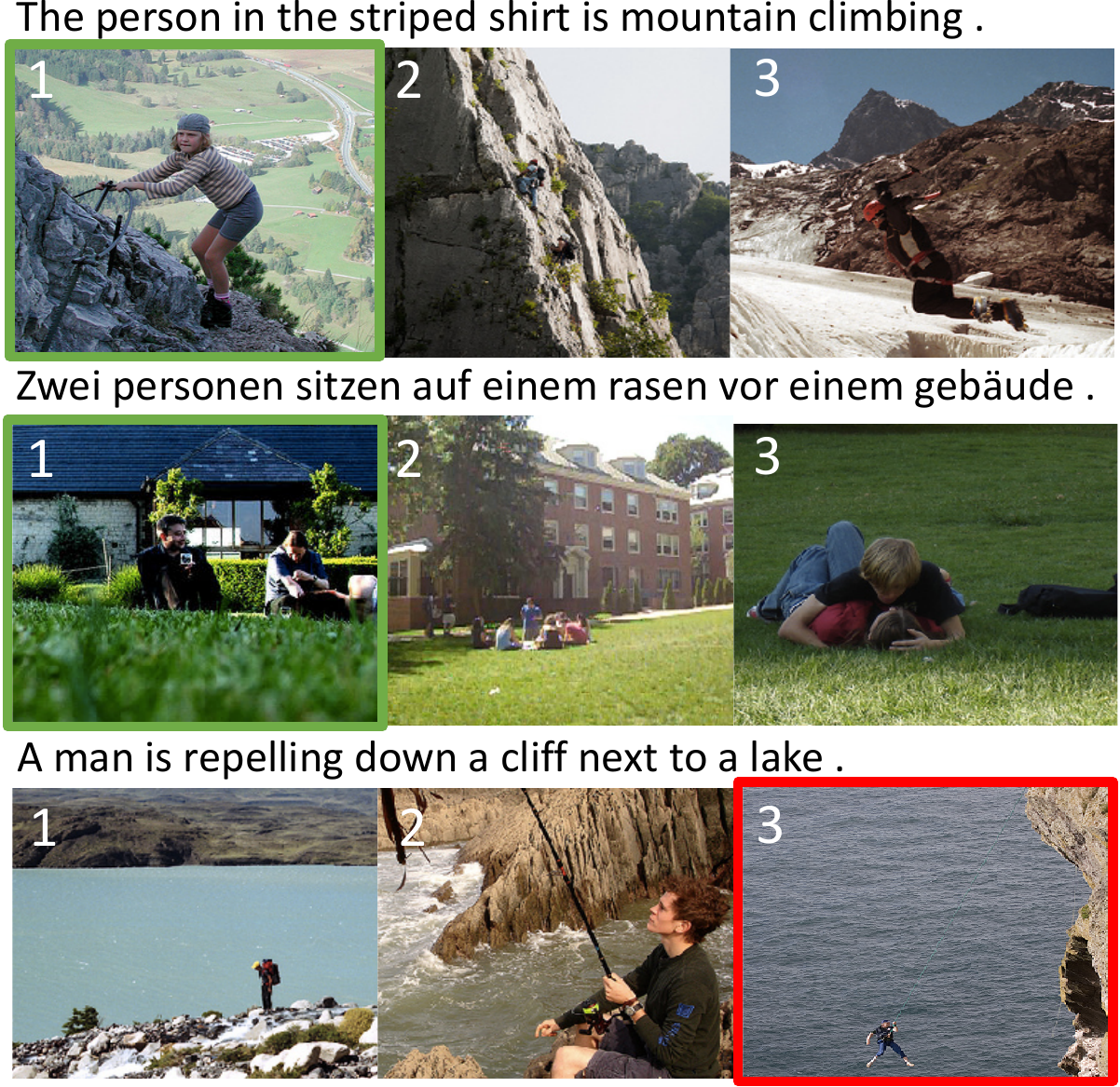}
    \caption{Qaulitative text-to-image matching results on Multi30K. Correct (colored in green) if ranked first.}\label{qual} 
    \vspace{-1.2em}
\end{figure}

\begin{figure}
    \centering
    \includegraphics[width=1.0\linewidth]{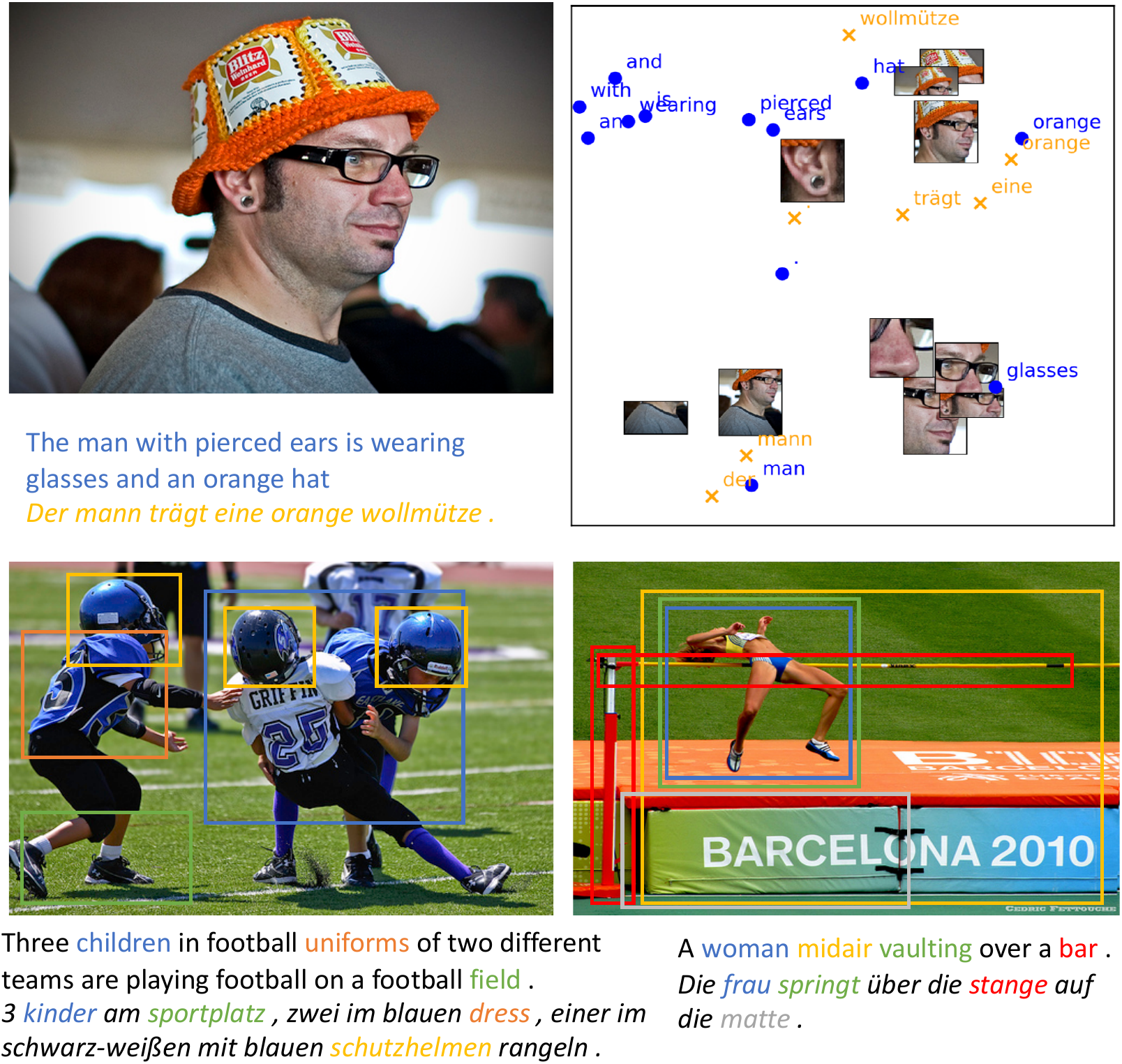}
    \caption{t-SNE visualization and grounding of the learned multilingual multimodal embeddings on Multi30K. Note the sentences are \textit{not} translation pairs.}\label{vis} 
    \vspace{-1.2em}
\end{figure}

\vspace{-0.2em}
\subsection{Qualitative Results and Grounding}
\vspace{-0.2em}
In Figure~\ref{qual} we samples some qualitative multilingual text-to-image matching results. In most cases our model successfully retrieve the one and only one correct image. 
Figure~\ref{vis} depicts the t-SNE visualization of the learned visually grounded multilingual embeddings of the $(v,e,g)$ pairs pivoted on $v$ in the Multi30K testing set. 
As evidenced, although the English and German sentences describe different aspects of the image, our model correctly aligns the shared semantics (\eg~(``\textit{man}'', ``\textit{mann}''), (``\textit{hat}'', ``\textit{wollmütze}'')) in the embedding space. 
Notably, the embeddings are visually-grounded as our model associate the multilingual phrases with exact visual objects (\eg~\textit{glasses} and \textit{ears}). 
We consider learning grounded multilingual multimodal dictionary as the promising next step.

As limitations we notice that actions and small objects are harder to align. Additionally, the alignments tends to be noun-phrase/object-based whereas spatial relationships  (\eg~``\textit{on}'', ``\textit{over}'') and quantifiers remain not well-aligned. Resolving these limitations will be our future work.

\vspace{-0.4em}
\section{Conclusion}
\vspace{-0.4em}
We have presented a novel VSE model facilitating multi-head attention with diversity to align different types of textual semantics and visual objects for learning grounded multilingual multimodal representations.
The proposed model obtains state-of-the-art results in the multilingual sentence-image matching task and the semantic textual similarity task on two benchmark datasets.

\vspace{-0.4em}
\section*{Acknowledgement}
\vspace{-0.4em}
This research is supported in part by the DARPA grants FA8750-18-2-0018 and FA8750-19-2-0501 under AIDA and LwLL program.
It is also supported by the IARPA grant via DOI/IBC number D17PC00340.
We would like to thank the anonymous reviewers for their constructive suggestions.

\bibliography{ref}
\bibliographystyle{acl_natbib}

\end{document}